\documentclass[10pt,twocolumn,letterpaper]{article}

\usepackage[pagenumbers]{cvpr} % To force page numbers, e.g. for an arXiv version
 \usepackage{booktabs}
 \usepackage{multirow}
 \usepackage{makecell}
 \usepackage{graphicx}
 \usepackage[table]{xcolor}

\definecolor{cvprblue}{rgb}{0.21,0.49,0.74}
\definecolor{mypink}{RGB}{219,112,147}
\usepackage[pagebackref,breaklinks,colorlinks,allcolors=cvprblue]{hyperref}

\def\confName{CVPR}
\def\confYear{2027}

\title{Event-Guided Stereo Video Super-Resolution}

\author{
Hanyu Xuan$^{1}$\quad Youtao Yao$^{1}$\quad Cheng Zhao$^{2}$\quad Kun Li$^{3}$\quad Xin Du$^{4}$\quad
Fei Wang$^{5}$\quad Zhiliang Wu$^{6}$\thanks{Corresponding author} \\~\\
{\fontsize{10.5pt}{12.5pt}\selectfont
$^{1}$Anhui University \quad
$^{2}$Shenzhen University \quad
$^{3}$United Arab Emirates University \quad
$^{4}$Zhejiang University \quad}\\
{\fontsize{10.5pt}{12.5pt}\selectfont
$^{5}$Institute of Artificial Intelligence, Hefei Comprehensive National Science Center \quad
$^{6}$Nanyang Technological University
}
}

\begin{document}
\maketitle

\begin{abstract}
Stereo Video Super-Resolution (SVSR) aims to reconstruct high-resolution stereo videos by exploiting temporal cues across consecutive frames and complementary spatial cues between stereo views. 
However, existing methods struggle to recover high-frequency texture details under rapid motion and complex illumination, compromising reconstruction quality and stereo consistency.
Event cameras provide valuable motion and texture cues with high temporal resolution and wide dynamic range, yet their potential for SVSR remains underexplored. 
To bridge this gap, we present the first investigation of event-guided SVSR and propose a novel framework.
%It jointly exploits event information, inter-frame temporal information, and complementary spatial information between stereo views for high-quality stereo video reconstruction. 
Specifically, we develop an Event-Guided Texture Enhancement (EGTE) module that combines bidirectional temporal feature propagation with event-guided texture refinement to restore high-frequency details in complex motion regions.
Subsequently, we introduce a Cross-View Texture Interaction (CVTI) module to model cross-view spatial correspondences and aggregate cross-view complementary texture information.
% , thereby improving stereo consistency.
With this design, our method effectively integrates 
event-driven motion and texture information, 
inter-frame temporal dependencies, 
and cross-view spatial relationships to 
recover motion-affected details and maintain stereo consistency.
% event-guided temporal texture enhancement with cross-view complementary information fusion to achieve high-quality and consistent stereo video reconstruction. 
Extensive experiments on both real-world and synthetic datasets demonstrate that our method consistently outperforms existing super-resolution methods in reconstruction quality and stereo consistency.
\textit{Project Page}: \href{https://wzlwlm13.github.io/EvSVSR/}{\textcolor{mypink}{https://EvSVSR.github.io}}.
\end{abstract}    
\section{Introduction}
\label{sec:intro}
Stereo Video Super-Resolution (SVSR)~\cite{xu2021stereo,9857357} aims to reconstruct high-resolution stereo videos from low-resolution stereo inputs, 
with promising applications in 3D vision, virtual reality, and autonomous driving.
Unlike conventional Video Super-Resolution (VSR) \cite{shang2024arbitrary,chen2026stcdit,10656731,11445549}, 
SVSR must exploit not only cross-frame temporal cues but also complementary cross-view information to recover fine-grained details while maintaining stereo consistency.
Existing SVSR methods~\cite{xu2021stereo,9857357} generally aggregate temporal information across successive frames and leverage spatial correspondences between views to improve reconstruction quality.

Despite these improvements, 
these methods rely heavily on RGB frames captured by conventional cameras. 
Nevertheless,
RGB cameras inherently suffer from limited temporal resolution arising from fixed sampling intervals.
Their finite exposure window further induces motion blur under rapid motion \cite{xiao2025event}.
Meanwhile,
complex illumination introduces noise and unpredictable intensity variations \cite{gallego2020event}.
Collectively,
these degradations impair motion estimation and hinder cross-view correspondence recovery, 
preventing the aggregation of complementary texture cues and degrading the reconstruction of fine-grained details.

\begin{figure}[!t]
%\vspace{-0.45cm}
\centering
\includegraphics[width=0.48\textwidth]{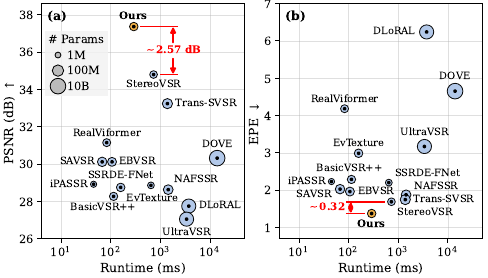}
\vspace{-0.6cm}
\caption{Reconstruction quality versus runtime on DSEC ~\cite{9387069},
where runtime is measured on $360\times270$ inputs per view.
(a) Average PSNR over the left and right views, measuring reconstruction fidelity. 
(b) EPE between stereo views, evaluating cross-view consistency. 
The circle size indicates the number of parameters on a logarithmic scale.
Our method achieves a favorable trade-off between reconstruction quality and computational efficiency.
Compared with the best-performing StereoVSR~\cite{xu2021stereo},
our method improves PSNR by $\sim 2.57$ dB and reduces EPE by $\sim 0.32$.
}
\label{fig:Intro}
\vspace{-0.45cm}
\end{figure}

Event cameras offer a promising alternative to mitigate such degradations. 
In contrast to RGB cameras, 
event cameras asynchronously capture brightness changes triggered by intensity variations.
Thanks to their high temporal resolution and wide dynamic range,
event signals provide abundant motion cues and high-frequency structural information \cite{xiao2026event,11444320}. 
Benefiting from these favorable properties, 
event signals have been extensively explored across a broad range of computer vision tasks \cite{10658213,11093954,11093199,11614017,11444320,11092933,10576683,10994378}.
In particular, 
recent studies \cite{10222922,kai2024evtexture,11369964,Kai_Zhang_Wang_Xiao_Xiong_Sun_2025,xiao2025event} 
further integrate event signals into VSR to boost the recovery of fine-grained details under rapid motion and complex illumination.
These studies verify the efficacy of event-based information for motion estimation and fine-detail reconstruction.

Despite these advances in event-guided VSR, 
existing methods~\cite{10222922,11369964,Kai_Zhang_Wang_Xiao_Xiong_Sun_2025,xiao2025event} target monocular video reconstruction.
In contrast to monocular counterparts, 
stereo videos contain inherent inter-view spatial correspondences 
alongside inter-frame temporal dependencies~\cite{10205204,Wu_Li_Xu_Fan_Yang_2026}.
Accordingly, event-guided SVSR requires joint exploitation of event-based information, inter-frame temporal dependencies, and cross-view complementary correspondences.
A straightforward adaptation of these event-guided VSR methods~\cite{11093954,xiao2026event,xiao2025event,Xiao_Wang_2026} to stereo inputs processes the left and right views individually.
However,
such view-wise separate processing fails to leverage inter-view complementary information and tends to yield inconsistent reconstructions between stereo views.

In this paper, we propose an Event-guided SVSR framework dubbed EvSVSR, 
which exploits event-driven motion information to recover motion-affected details, 
utilizes inter-frame temporal dependencies to capture frame-wise dynamics, 
and leverages cross-view spatial relationships to enforce stereo consistency. 
Specifically, EvSVSR consists of two core modules: 
Event-Guided Texture Enhancement (EGTE) and 
Cross-View Texture Interaction (CVTI). 
EGTE performs bidirectional recurrent propagation 
to align intra-view temporal features to the current frame.
Within this propagation, 
event features guide the recurrent state updates
to refine frame-level texture features and restore high-frequency details in complex motion regions.
Subsequently, CVTI enables bidirectional feature interaction across stereo views to establish cross-view spatial correspondences.

Extensive experiments on the real-world DSEC \cite{9387069} and synthetic SVSR-Set \cite{9857357} datasets demonstrate that 
EvSVSR consistently outperforms existing methods in reconstruction quality and stereo consistency, as shown in Fig.~\ref{fig:Intro}.
The main contributions of our work are summarized as follows:
\begin{itemize}
\item We investigate the potential of event information for SVSR  and formulate the event-guided SVSR task. 
To the best of our knowledge, this is the first attempt to exploit event information for boosting SVSR performance.

\item We develop an event-guided SVSR framework,
which effectively combines event-driven motion cues and cross-view complementary spatial information to achieve high-quality stereo video reconstruction while maintaining stereo correspondence consistency.

\item Extensive experiments on both real-world and synthetic stereo video datasets verify the superior quantitative and qualitative performance of our method. 
Notably, 
% our method provides insights for future research on event-guided SVSR.
our empirical findings provide practical references for the design of subsequent event-guided SVSR frameworks.

\end{itemize}
\section{Related Work}
\label{Related Work}

\noindent\textbf{1) Video Super-Resolution}
% \vspace{0.1cm}

Early VSR methods~\cite{9025464,10462902,wen2022video,9879776} typically employed sliding-window architectures,
which reconstruct target frames by aligning and aggregating features from adjacent frames.
However, their fixed temporal windows restrict long-range information propagation and incur redundant computations among overlapping frames~\cite{9879547,11445549}. 
To address these limitations, 
recurrent VSR methods, 
including BasicVSR~\cite{9577681} and BasicVSR++~\cite{9879156}, propagate and update features frame by frame to efficiently capture long-range temporal dependencies.
Transformer-based methods~\cite{10656731,shang2024arbitrary} have been developed to model long-range spatiotemporal dependencies, while diffusion-based methods~\cite{chen2026stcdit,wang2025seedvr} leverage generative priors to enhance fine-grained details and improve perceptual quality.
Nevertheless, these methods rely primarily on frame-only visual cues and degrade under complex motion.

\vspace{0.1cm}
\noindent\textbf{2) Event-Guided Video Super-Resolution}
\vspace{0.1cm}

To address this challenge, 
recent studies have explored event-based information to provide valuable motion cues for VSR.
For example,
EBVSR~\cite{10222922} introduced event signals to estimate motion and perform temporal alignment.
EvTexture~\cite{kai2024evtexture} refined texture details with event cues, while EvTexture++~\cite{11369964} further improved temporal consistency by leveraging event cues for frame alignment.
E2VSR~\cite{xiao2026event} exploited event-confidence adaptive propagation for efficient online reconstruction, 
while BluR-EVSR~\cite{Xiao_Wang_2026} combined blurry frames with event-guided motion cues for video restoration. 
These existing event-guided VSR methods verify the efficacy of event-based information for video reconstruction.
However, these methods are mainly designed for monocular videos and ignore spatial correspondences and complementary cues across stereo views.

\begin{figure*}[!t]
\centering
\includegraphics[width=0.95\textwidth]{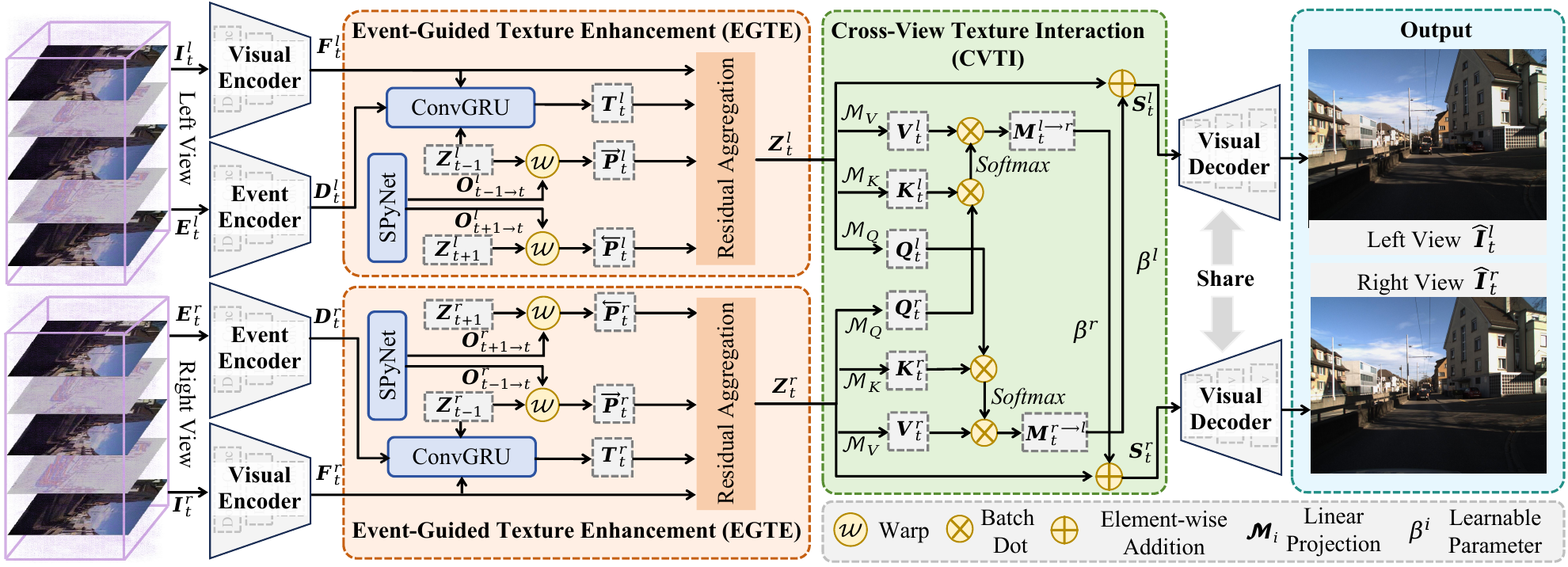}
\vspace{-0.3cm}
\caption{\textbf{Overview of the proposed Event-guided Stereo Video Super-Resolution (EvSVSR) framework.} 
Our EvSVSR consists of two core modules: 
Event-Guided Texture Enhancement (EGTE) and 
Cross-View Texture Interaction (CVTI).
The EGTE module performs bidirectional recurrent propagation to align intra-view temporal features to the current frame,
while the CVTI module establishes cross-view spatial correspondences to facilitate bidirectional feature interaction across stereo views.
The pre-trained SPyNet~\cite{8099774} is adopted for optical flow estimation.
}
\label{fig:framework}
\vspace{-0.3cm}
\end{figure*}

\vspace{0.1cm}
\noindent\textbf{3) Event-based Computer Vision}
\vspace{0.1cm}

Event cameras~\cite{gallego2020event,xiao2025event} provide high temporal resolution and high dynamic range, enabling them to capture rapid brightness changes,
which are hardly visible in RGB frames. 
Event cues have been explored in various computer vision tasks, including image deblurring~\cite{10658213,11093954}, video frame interpolation~\cite{11093199,11614017}, optical flow estimation~\cite{11444320,11092933}, and image super-resolution~\cite{10576683,10994378}. 
These studies demonstrate that event signals provide strong motion cues,
which facilitate motion estimation and visual reconstruction.

\vspace{0.1cm}
\noindent\textbf{4) Stereo Video Super-Resolution}
\vspace{0.1cm}

Unlike conventional VSR~\cite{zhuang2026flashvsr,zhu2026fcvsr,sun2026one,NEURIPS2025_7b04ec5f,11417318} and 
Stereo Image Super-Resolution (SISR)~\cite{dai2021feedback,9522905,9857435}, 
SVSR needs to exploit 
not only inter-frame temporal dependencies, but also
spatial correspondences across stereo views.
For example, StereoVSR~\cite{xu2021stereo} exploited inter-frame temporal dependencies and cross-view spatial correlations via joint feature aggregation to boost reconstruction quality.
E-SEVSR~\cite{10440343} introduced a stereo endoscopic attention to exploit cross-view cues for endoscopic SVSR.
Trans-SVSR~\cite{9857357} adopted parallax-aware spatiotemporal attention to model cross-view correspondences and temporal dependencies.

Despite achieving promising results, 
these SVSR methods~\cite{9857357,10440343,xu2021stereo} still encounter difficulties under rapid motion and severe blurring. 
These degradations hinder temporal feature alignment and cross-view correspondence modeling, leading to degraded reconstruction quality and stereo consistency.
To this end, we propose an event-guided SVSR framework that exploits event-driven motion cues to strengthen cross-view consistency and intra-view temporal coherence.

\section{Methodology}
\label{Methodology}

\subsection{Problem Formulation}
\label{Overview}

Let $v\in\{l,r\}$ index the left and right views of a stereo pair.
We denote the low-resolution video sequence of view $v$ as
$\boldsymbol{\mathcal{I}}^{v}=\{\textbf{\emph{I}}_t^v\}_{t=1}^T$,
where $\textbf{\emph{I}}^{v}_{t} \in \mathbb{R}^{h \times w \times 3}$ corresponds to the $t$-th frame.
For each view,
its paired event stream is
$\boldsymbol{\mathcal{E}}^{v}=
\{\textbf{\emph{E}}_t^v\}_{t=2}^{T}$, 
where $\textbf{\emph{E}}^{v}_{t}$ accumulates events triggered between adjacent frames $\textbf{\emph{I}}^{v}_{t-1}$ and $\textbf{\emph{I}}^{v}_{t}$.
Given these low-resolution frames and their accompanying event streams,
the goal of event-guided SVSR is to reconstruct high-resolution stereo video sequences $\widehat{\boldsymbol{\mathcal{I}}}^{v}=\{\widehat{\textbf{\emph{I}}}_t^v\}_{t=1}^T$.
The reconstructed outputs should preserve spatial fidelity, temporal smoothness, and stereo consistency against the ground-truth high-resolution
sequences $\boldsymbol{\mathcal{H}}^{v}=\{\textbf{\emph{H}}_t^v\}_{t=1}^T$.
Here, $\widehat{\textbf{\emph{I}}}_t^v, \textbf{\emph{H}}_t^v\in \mathbb{R}^{sh \times sw \times 3}$, and $s$ is the scaling factor for super-resolution.

To achieve this goal,
we design an EvSVSR framework (Fig.~\ref{fig:framework}),
which simultaneously models event-driven motion information, 
inter-frame temporal dependencies, 
and cross-view spatial relationships to 
recover motion-affected details and maintain stereo consistency.
Specifically,
we adopt a visual encoder to extract visual features
$\textbf{F}_t^v$
from the left- and right-view frames.
The paired inter-frame events are 
% converted into voxel grids and
fed into an event encoder to yield event features
$\textbf{D}_t^v$.
To enable bidirectional temporal propagation, we estimate the forward optical flow
$\textbf{\emph{O}}_{t-1\rightarrow{t}}^{v}$ between adjacent frames
$\textbf{\emph{I}}_{t}^v$ and $\textbf{\emph{I}}_{t-1}^v$.
The backward branch follows the same procedure in the reverse temporal direction and yields $\textbf{\emph{O}}_{t+1\rightarrow{t}}^{v}$.

\subsection{Event-Guided Texture Enhancement}
Existing SVSR methods~\cite{9857357,10440343,xu2021stereo}  rely primarily on RGB inputs to reconstruct high-resolution stereo frames.
However, rapid motion and blurring frequently obscure high-frequency texture details and degrade reconstruction accuracy.
As an alternative,
event signals provide complementary information with high temporal resolution, facilitating texture restoration within complex-motion regions~\cite{gallego2020event,xiao2025event}.

% 不知道是否准确
A naive adaptation of event-guided VSR methods 
\cite{xiao2025event,Kai_Zhang_Wang_Xiao_Xiong_Sun_2025,kai2024evtexture,xiao2026event}
to stereo inputs processes the left and right views separately, which leads to suboptimal reconstruction.
On the one hand,
this solution fails to exploit cross-view complementary cues and tends to produce inconsistent reconstructions between stereo views.
On the other hand,
these methods decouple motion and texture during temporal feature propagation. 
In turn, warping errors accumulate across consecutive frames, 
yielding blurry dynamic textures and further degrading stereo consistency.

Specifically, the estimated forward and backward optical flows 
$\textbf{\emph{O}}_{t-1\rightarrow{t}}^{v}$ and $\textbf{\emph{O}}_{t+1\rightarrow{t}}^{v}$
are adopted to warp the hidden states $\textbf{\emph{Z}}_{t-1}^v$ and $\textbf{\emph{Z}}_{t+1}^v$ 
from frames $\textbf{\emph{I}}_{t-1}^v$ and $\textbf{\emph{I}}_{t+1}^v$ to the current frame $\textbf{\emph{I}}_{t}^v$, respectively.
The forward and backward motion-aligned features 
$\overrightarrow{\textbf{\emph{P}}_{{t}}^{v}}$ and 
$\overleftarrow{\textbf{\emph{P}}_{{t}}^{v}}$
at the current frame $\textbf{\emph{I}}_{t}^v$ are defined as:
\begin{equation}
\overrightarrow{\textbf{\emph{P}}_{{t}}^{v}}=\mathcal{W}(\textbf{\emph{Z}}_{t-1}^v, \textbf{\emph{O}}_{t-1\rightarrow{t}}^{v}),
\quad
\overleftarrow{\textbf{\emph{P}}_{{t}}^{v}}=\mathcal{W}(\textbf{\emph{Z}}_{t+1}^v, \textbf{\emph{O}}_{t+1\rightarrow{t}}^{v}),
\label{S2}
\end{equation}
where 
$\mathcal{W}(\cdot,\cdot)$ denotes the feature warping operator.
The hidden state \(\textbf{\emph{Z}}_{0}^v\) is initialized to zeros.

Built upon these aligned features,
we introduce a ConvGRU layer~\cite{ballas2015delving},
denoted by $\mathcal{C}(\cdot,\cdot,\cdot)$,
to adaptively integrate 
the previous hidden state $\textbf{\emph{Z}}_{t-1}^{v}$ 
with the visual features $\textbf{\emph{F}}_{t}^{v}$ 
and event features $\textbf{\emph{D}}_{t}^{v}$ at the current frame:
\begin{equation}
\textbf{\emph{T}}_{{t}}^{v}=\mathcal{C}(\textbf{\emph{Z}}_{{t-1}}^{v}, \textbf{\emph{F}}_{t}^v, \textbf{\emph{D}}_{t}^v).
\label{S3}
\end{equation}
Event information delivers complementary high-frequency details that cannot be reliably recovered from motion-blurred visual frames. 
Integrating motion and event cues
into recurrent propagation enables continuous enhancement,
which strengthens detail preservation in complex-motion regions.
The enhanced features are then fused via residual blocks $\mathcal{R}(\cdot)$
to update the hidden state:
\begin{equation}
\textbf{\emph{Z}}_{{t}}^{v}=\mathcal{R}(\overrightarrow{\textbf{\emph{P}}_{{t}}^{v}}, \textbf{\emph{F}}_{t}^v, \overleftarrow{\textbf{\emph{P}}_{{t}}^{v}}, \textbf{\emph{T}}_{t}^v),
\label{S4}
\end{equation}
In this way, the EGTE module aggregates temporal information from both preceding and subsequent frames.

\subsection{Cross-View Texture Interaction}
Although the EGTE module enhances  temporal texture features
via intra-view temporal propagation, 
directly reconstructing of high-resolution frames without cross-view interaction tends to produce misaligned textures between stereo views.
To address this issue, we introduce a CVTI to model cross-view spatial correspondences and enable cross-view texture communication.

Specifically, 
given the event-enhanced texture feature $\textbf{\emph{Z}}_{{t}}^v$, 
we apply linear projections $\mathcal{M}_{Q}(\cdot)$, $\mathcal{M}_{K}(\cdot)$ and $\mathcal{M}_{V}(\cdot)$ to derive the intra-view query $\textbf{\emph{Q}}$, key $\textbf{\emph{K}}$, and value $\textbf{\emph{V}}$:
\begin{equation}
\textbf{\emph{Q}}_t^v, ~\textbf{\emph{K}}_t^v, ~\textbf{\emph{V}}_t^v =\mathcal{M}_Q(\textbf{\emph{Z}}_{{t}}^v), ~\mathcal{M}_K(\textbf{\emph{Z}}_{{t}}^v), ~\mathcal{M}_V(\textbf{\emph{Z}}_{{t}}^v).
\label{S4}
\end{equation}
We then compute bidirectional attention weights 
using the query $\textbf{\emph{Q}}$ from one view and the key $\textbf{\emph{K}}$ from the other view.
These attention weights aggregate complementary texture cues from the opposite view.
The resulting cross-view texture features are given by:
\begin{equation}
\begin{split}
&\textbf{\emph{M}}_t^{r\rightarrow{l}}=\sigma(\frac{\textbf{\emph{Q}}_t^l\cdot{\textbf{\emph{K}}_t^r}^T}{\sqrt{c}})\cdot\textbf{\emph{V}}_t^r, \\
&\textbf{\emph{M}}_t^{l\rightarrow{r}}=\sigma(\frac{\textbf{\emph{Q}}_t^r\cdot{\textbf{\emph{K}}_t^l}^T}{\sqrt{c}})\cdot\textbf{\emph{V}}_t^l,
\label{S}
\end{split}
\end{equation}
where $\sigma(\cdot)$ indicates softmax function.
This mechanism captures stereo spatial correspondences and facilitates the exchange of cross-view complementary texture cues.

\begin{table*}[t]
\centering
\footnotesize
\renewcommand\tabcolsep{4pt}
\caption{Quantitative comparison of representative methods 
\cite{9879156,Li_Liu_Shang_Liu_Wan_Feng_2024,sun2026one,10.1007/978-3-031-73397-0_24,liu2025ultravsr,NEURIPS2025_7b04ec5f,kai2024evtexture,10222922,dai2021feedback,9522905,9857435,xu2021stereo,9857357}
for $\times 4$ stereo video super-resolution on DSEC and SVSR-Set datasets. $\uparrow$/$\downarrow$ indicates that higher/lower values are better. 
The best results are highlighted in \textbf{bold}.}
 \vspace{-0.3cm}
\label{tab:main_results}
%\resizebox{\textwidth}{!}{%
\begin{tabular}{cl cccc cccc cccc c}
\toprule
\multicolumn{2}{c}{\multirow{2}{*}{\textbf{Method}}} 
& \multicolumn{4}{c}{\textbf{Left view}} 
& \multicolumn{4}{c}{\textbf{Right view}} 
& \multicolumn{4}{c}{\textbf{Average}} 
& \multirow{2}{*}{\textit{EPE}$\downarrow$} \\
\cmidrule(lr){3-6} \cmidrule(lr){7-10} \cmidrule(lr){11-14}
\multicolumn{2}{c}{} 
& \textit{PSNR}$\uparrow$ & \textit{SSIM}$\uparrow$ & \textit{LPIPS}$\downarrow$ & $E_{warp}\downarrow$
& \textit{PSNR}$\uparrow$ & \textit{SSIM}$\uparrow$ & \textit{LPIPS}$\downarrow$ & $E_{warp}\downarrow$
& \textit{PSNR}$\uparrow$ & \textit{SSIM}$\uparrow$ & \textit{LPIPS}$\downarrow$ & $E_{warp}\downarrow$
& \\
\midrule
% ===================== AAA Dataset =====================
\multicolumn{15}{c}{\textit{DSEC Dataset}} \\
\midrule
\multirow{6}{*}{\textbf{i)}}
& BasicVSR++   & 27.84 & 0.831 & 0.385 & 0.044 & 28.70 & 0.841 & 0.379 & 0.032 & 28.27 & 0.836 & 0.382 & 0.038 & 2.289 \\
& SAVSR        & 29.73 & 0.882 & 0.366 & 0.039 & 30.50 & 0.889 & 0.360 & 0.028 & 30.12 & 0.885 & 0.363 & 0.034 & 2.029 \\
& DLoRAL       & 27.60 & 0.787 & 0.371 & 0.032 & 27.90 & 0.788 & 0.373 & 0.030 & 27.75 & 0.788 & 0.372 & 0.031 & 6.241 \\
& RealViformer & 30.73 & 0.880 & 0.305 & 0.013 & 31.56 & 0.886 & 0.303 & 0.011 & 31.15 & 0.883 & 0.304 & 0.012 & 4.184 \\
& UltraVSR     & 27.05 & 0.804 & 0.354 & 0.040 & 27.19 & 0.805 & 0.353 & 0.038 & 27.12 & 0.805 & 0.353 & 0.039 & 3.172 \\
& DOVE         & 30.03 & 0.864 & 0.332 & 0.013 & 30.61 & 0.871 & 0.329 & 0.012 & 30.32 & 0.867 & 0.331 & 0.013 & 4.655 \\
\midrule
\multirow{2}{*}{\textbf{ii)}}
& EvTexture    & 28.24 & 0.844 & 0.390 & 0.051 & 29.25 & 0.855 & 0.384 & 0.038 & 28.75 & 0.849 & 0.387 & 0.045 & 2.990 \\
& EBVSR        & 29.54 & 0.871 & 0.364 & 0.037 & 30.69 & 0.881 & 0.359 & 0.026 & 30.12 & 0.876 & 0.362 & 0.032 & 1.964 \\
\midrule
\multirow{3}{*}{\textbf{iii)}}
& SSRDE-FNet   & 28.40 & 0.864 & 0.372 & 0.050 & 29.31 & 0.873 & 0.366 & 0.037 & 28.86 & 0.868 & 0.369 & 0.044 & 2.204 \\
& iPASSR       & 28.50 & 0.862 & 0.369 & 0.048 & 29.33 & 0.870 & 0.363 & 0.036 & 28.92 & 0.866 & 0.366 & 0.042 & 2.233 \\
& NAFSSR       & 28.21 & 0.890 & 0.360 & 0.048 & 29.05 & 0.897 & 0.354 & 0.036 & 28.63 & 0.893 & 0.357 & 0.042 & 1.877 \\
\midrule
\multirow{2}{*}{\textbf{iv)}}
& StereoVSR   & 34.25 & 0.921 & 0.313 & 0.010 & 35.34 & 0.928 & 0.310 & 0.008 & 34.80 & 0.925 & 0.312 & 0.009 & 1.691 \\
&Trans-SVSR   & 32.68 & 0.904 & 0.354 & 0.015 & 33.79 & 0.912 & 0.349 & 0.011 & 33.24 & 0.908 & 0.352 & 0.013 & 1.749  \\
\midrule
\multicolumn{2}{c}{\cellcolor{blue!10}\textbf{Ours}} 
& \cellcolor{blue!10}\textbf{36.78} & \cellcolor{blue!10}\textbf{0.944} &\cellcolor{blue!10} \textbf{0.260} &\cellcolor{blue!10} \textbf{0.008} 
&\cellcolor{blue!10} \textbf{37.96} &\cellcolor{blue!10} \textbf{0.949} & \cellcolor{blue!10}\textbf{0.256} &\cellcolor{blue!10}\textbf{0.006} 
&\cellcolor{blue!10} \textbf{37.37} & \cellcolor{blue!10}\textbf{0.947} & \cellcolor{blue!10}\textbf{0.258} & \cellcolor{blue!10}\textbf{0.007} 
&\cellcolor{blue!10} \textbf{1.376} \\
\midrule
% ===================== BBB Dataset =====================
\multicolumn{15}{c}{\textit{SVSR-Set Dataset}} \\
\midrule
\multirow{6}{*}{\textbf{i)}}
& BasicVSR++   & 27.38 & 0.876 & 0.254 & 0.006 & 26.75 & 0.868 & 0.260 & 0.006 & 27.07 & 0.872 & 0.257 & 0.006 & 2.649 \\
& SAVSR        & 27.49 & 0.880 & 0.241 & 0.005 & 26.91 & 0.873 & 0.246 & 0.006 & 27.20 & 0.877 & 0.244 & 0.006 & 2.624 \\
& DLoRAL       & 27.23 & 0.809 & 0.390 & 0.006 & 26.89 & 0.810 & 0.388 & 0.006 & 27.06 & 0.809 & 0.389 & 0.006 & 6.465 \\
& RealViformer & 29.72 & 0.891 & 0.280 & \textbf{0.001} & 29.10 & 0.886 & 0.285 & 0.002 & 29.41 & 0.888 & 0.282 & 0.002 & 4.410 \\
& UltraVSR     & 25.65 & 0.817 & 0.326 & 0.008 & 25.38 & 0.816 & 0.327 & 0.008 & 25.52 & 0.817 & 0.327 & 0.008 & 4.073 \\
& DOVE         & 29.65 & 0.884 & 0.275 & 0.002 & 29.13 & 0.882 & 0.275 & 0.002 & 29.39 & 0.883 & 0.275 & 0.002 & 4.974 \\
\midrule
\multirow{2}{*}{\textbf{ii)}}
& EvTexture    & 27.17 & 0.878 & 0.251 & 0.006 & 26.30 & 0.861 & 0.258 & 0.008 & 26.74 & 0.869 & 0.255 & 0.007 & 3.342 \\
& EBVSR        & 27.91 & 0.882 & 0.265 & 0.004 & 27.30 & 0.875 & 0.271 & 0.005 & 27.61 & 0.879 & 0.268 & 0.005 & 2.645 \\
\midrule
\multirow{3}{*}{\textbf{iii)}}
& SSRDE-FNet   & 26.82 & 0.875 & 0.262 & 0.007 & 26.34 & 0.870 & 0.266 & 0.008 & 26.58 & 0.872 & 0.264 & 0.008 & 2.697 \\
& iPASSR       & 27.04 & 0.877 & 0.261 & 0.007 & 26.52 & 0.872 & 0.264 & 0.007 & 26.78 & 0.874 & 0.262 & 0.007 & 2.707 \\
& NAFSSR       & 26.61 & 0.865 & 0.258 & 0.008 & 26.06 & 0.858 & 0.261 & 0.009 & 26.34 & 0.861 & 0.260 & 0.009 & 2.635 \\
\midrule
\multirow{2}{*}{\textbf{iv)}}
& StereoVSR   & 31.82      &  0.926     & 0.220      & 0.014      & 31.23      &  0.921     & 0.223      &  0.015     & 31.53      & 0.924      & 0.222      & 0.015      &  2.407      \\
&Trans-SVSR   & 30.38      &  0.906     & 0.257      &   0.018    &    29.71   &  0.899     &  0.263     &  0.020     &  30.05     & 0.903      &  0.260     &  0.019     &  2.532      \\
\midrule
\multicolumn{2}{c}{\cellcolor{blue!10}\textbf{Ours}} 
&\cellcolor{blue!10} \textbf{33.84} &\cellcolor{blue!10} \textbf{0.945} & \cellcolor{blue!10}\textbf{0.175} & \cellcolor{blue!10}\textbf{0.001} 
& \cellcolor{blue!10}\textbf{33.37} &\cellcolor{blue!10} \textbf{0.943} & \cellcolor{blue!10}\textbf{0.176} & \cellcolor{blue!10}\textbf{0.001} 
& \cellcolor{blue!10}\textbf{33.61} & \cellcolor{blue!10}\textbf{0.944} &\cellcolor{blue!10} \textbf{0.176} &\cellcolor{blue!10} \textbf{0.001} 
&\cellcolor{blue!10} \textbf{1.994} \\
\bottomrule
\end{tabular}%
%}
 \vspace{-0.3cm}
\end{table*}

The resulting cross-view features 
$\textbf{\emph{M}}_t^{r\rightarrow{l}}$ and
$\textbf{\emph{M}}_t^{l\rightarrow{r}}$
are combined with the original event-enhanced features 
$\textbf{\emph{Z}}_{{t}}^l$ and
$\textbf{\emph{Z}}_{{t}}^r$
via residual connections:
\begin{equation}
\textbf{\emph{S}}_t^{l}=\beta^l\textbf{\emph{M}}_t^{r\rightarrow{l}}+\textbf{\emph{Z}}_{{t}}^l,
~~~~\textbf{\emph{S}}_t^{r}=\beta^r\textbf{\emph{M}}_t^{l\rightarrow{r}}+\textbf{\emph{Z}}_{{t}}^r,
\label{S1}
\end{equation}
where $\beta^l$ and $\beta^r$ are learnable scaling parameters that modulate the contribution of complementary cues transferred from the opposite view.
Subsequently,
the combined features $\textbf{\emph{S}}_t^{l}$ and $\textbf{\emph{S}}_t^{r}$ are fed into a shared decoder to generate high-resolution left- and right-view frames $\widehat{\textbf{\emph{I}}}_t^l$ and $\widehat{\textbf{\emph{I}}}_t^r$. 
This design yields high-quality stereo frames with faithful high-frequency details and consistent stereo perception.

\subsection{Loss Functions}

We employ the Charbonnier loss~\cite{8100101} to supervise the reconstruction of both left and right views.
The reconstruction loss is formulated as:
\begin{equation}
\mathcal{L}=\frac{1}{2T}\sum_{v\in\{l,r\}}\sum_{t=1}^{T}\sqrt{\left\| \widehat{\textbf{\emph{I}}}_t^v - \textbf{\emph{H}}_t^v \right\|^2 + \varepsilon},
\label{eq:3d}
\end{equation}
where $\varepsilon=1 \times 10^{-6}$ is set for numerical stability.
\section{Experiments}
\subsection{Experimental Setting}

\noindent\textbf{Datasets.}
We conduct extensive experiments on one synthetic dataset and one real-world dataset to validate the proposed framework.

\emph{\textbf{Synthetic Dataset:}} 
The SVSR-Set dataset~\cite{9857357} is tailored for stereo VSR and consists of 71 high-definition stereo video sequences with diverse scenes and motion patterns.
Following the standard VSR setting \cite{Kai_Zhang_Wang_Xiao_Xiong_Sun_2025,xiao2025event,11369964}, 
we generate low-resolution stereo inputs via bicubic downsampling of the original high-resolution videos.
Since SVSR-Set does not provide native event streams, 
we employ ESIM~\cite{pmlr-v87-rebecq18a} as an event simulator to synthesize event streams for the left and right views separately.

\emph{\textbf{Real-World Dataset:}} 
The DSEC dataset~\cite{9387069} comprises $53$ sequences collected in driving scenarios with diverse illumination conditions.
It provides temporally synchronized stereo RGB videos and corresponding event streams, captured by paired RGB–event cameras.
The RGB sequences are captured at $1440\times1080$ with a frame rate of $20$ FPS, while the event streams have a spatial resolution of $640\times480$.

\noindent\textbf{Implementation Details.}
We implement the proposed framework in PyTorch and take five consecutive stereo frame pairs as input.
The visual encoder employs a $3\times3$ convolutional layer followed by eight residual blocks. 
The event encoder adopts a five-level U-Net~\cite{ronneberger2015u} with encoder channel numbers of $16$, $32$, $64$, $128$, and $128$.
The visual decoder up-samples the propagated features by a factor of $4\times$ via two successive stages, each consisting of a convolution and PixelShuffle~\cite{shi2016real}. 
It then predicts an RGB residual and adds it to the bi-linearly up-sampled low-resolution frame.

For training, the batch size is set to $5$,
and the spatial size of low-resolution input patches is  $64\times64$. 
Random horizontal and vertical flips are adopted for data augmentation. 
Inter-frame events from both views are converted into voxel grids with $15$ temporal bins. 
The resulting voxel grids share the same spatial resolution as the low-resolution input frames. 
The Adam optimizer, with momentum parameters $\beta_1=0.9$ and $\beta_2=0.99$,
is adopted,
alongside a cosine annealing learning rate schedule. 
For DSEC~\cite{9387069},
training proceeds for $300$K iterations with an initial learning rate of $2\times10^{-4}$. 
For SVSR-Set~\cite{9857357}, 
the DSEC-pretrained model is fine-tuned for $40$K iterations with an initial learning rate of $5\times10^{-5}$.

\begin{table*}[t]
\centering
\renewcommand\tabcolsep{1.8pt}
\scriptsize
\caption{Computational efficiency comparison with existing super-resolution methods \cite{9879156,Li_Liu_Shang_Liu_Wan_Feng_2024,sun2026one,10.1007/978-3-031-73397-0_24,liu2025ultravsr,NEURIPS2025_7b04ec5f,kai2024evtexture,10222922,dai2021feedback,9522905,9857435,xu2021stereo,9857357}.}
% \vspace{-0.3cm}
\label{tab:efficiency}
\begin{tabular}{lcccccccccccccc}
\toprule
\textbf{Method} & BasicVSR++ & SAVSR & DLoRAL & RealViformer & UltraVSR & DOVE & EvTexture & EBVSR & SSRDE-FNet & iPASSR & NAFSSR & StereoVSR &Trans-SVSR  & \textbf{Ours} \\
\midrule
\textbf{Params} (\textit{M}) & 7.32 & 18.89 & 1771.64 & 5.82 & 1771.36 & 10548.57 & 8.90 & 12.15 & 2.26 & 1.43 & 23.83 &4.61    &33.80   & 8.72 \\
\textbf{Runtime} (\textit{ms}) & 113.8 & 67.0 & 3723.4 & 82.6 & 3361.6 & 13909.0 & 158.2 & 105.8 & 643.7 & 45.0 & 1434.7 &726.6   &1382.5 & 290.4 \\
\bottomrule
\end{tabular}
% \vspace{-0.1cm}
\end{table*}

\begin{figure*}[!t]
\centering
%\vspace{-0.4cm}
\includegraphics[width=1\textwidth]{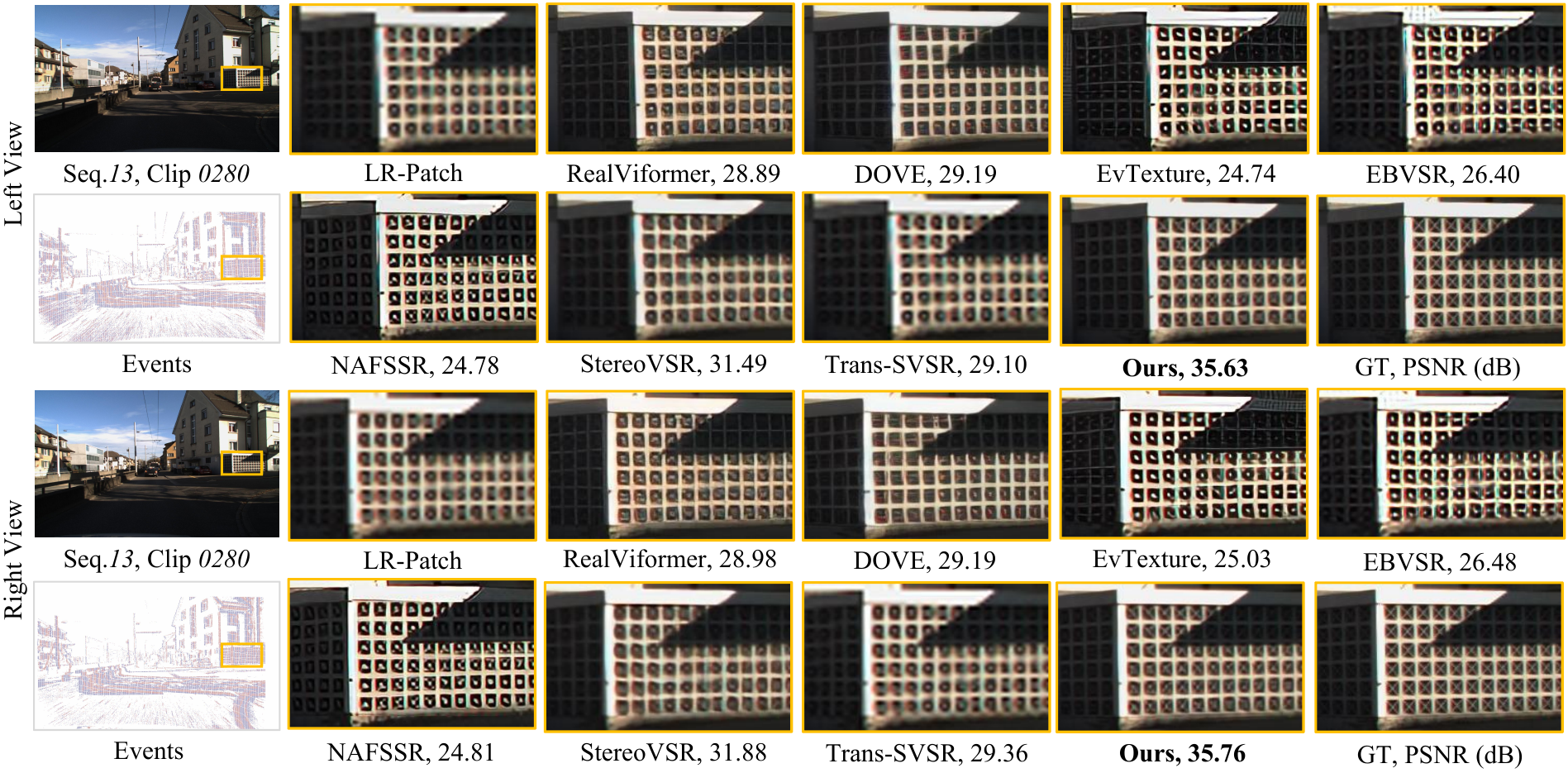}
% \vspace{-0.7cm}
\caption{Qualitative comparison of existing methods \cite{10.1007/978-3-031-73397-0_24, NEURIPS2025_7b04ec5f, kai2024evtexture, 10222922, 9857435, xu2021stereo, 9857357} for $\times 4$ stereo video super-resolution on the real-world DSEC. Best viewed at $400$\% zoom.
}
\label{fig:DSEC}
% \vspace{-0.25cm}
\end{figure*}

\noindent\textbf{Baselines.}
To comprehensively evaluate the proposed method, 
we compare it against thirteen representative super-resolution methods from four categories.
\begin{enumerate}[label=\roman*)]
\item  \textbf{RGB-based VSR methods:} BasicVSR++~\cite{9879156}, SAVSR~\cite{Li_Liu_Shang_Liu_Wan_Feng_2024}, DLoRAL~\cite{sun2026one}, RealViformer~\cite{10.1007/978-3-031-73397-0_24}, UltraVSR~\cite{liu2025ultravsr}, and DOVE~\cite{NEURIPS2025_7b04ec5f}. 
\item \textbf{Event-guided VSR methods:} EvTexture~\cite{kai2024evtexture} and EBVSR~\cite{10222922}.
\item \textbf{RGB-based SISR methods:} SSRDE-FNet~\cite{dai2021feedback}, iPASSR~\cite{9522905}, and NAFSSR~\cite{9857435}. 
\item  \textbf{RGB-based SVSR methods:} StereoVSR~\cite{xu2021stereo} and Trans-SVSR~\cite{9857357}. 
\end{enumerate}
For a fair comparison, 
these baselines are evaluated on the same test sets using consistent evaluation metrics.

\noindent\textbf{Evaluation Metrics.}
We adopt five metrics to evaluate reconstruction quality, temporal consistency, and geometric accuracy: PSNR~\cite{11092792}, SSIM~\cite{1284395}, LPIPS~\cite{8578166}, flow warping error ($E_{\mathrm{warp}}$)~\cite{Wu_Li_Xu_Fan_Yang_2026}, and end-point error (EPE)~\cite{10205204}.
Specifically, PSNR measures pixel-level fidelity, SSIM evaluates structural similarity, and LPIPS assesses perceptual similarity.
$E_{\mathrm{warp}}$ measures temporal consistency by evaluating the discrepancy between motion-aligned consecutive frames.
EPE evaluates geometric accuracy by measuring the average absolute error between estimated and ground-truth disparity maps.
Higher PSNR and SSIM, together with lower LPIPS, $E_{\mathrm{warp}}$ and EPE, indicate better performance.

\begin{figure*}[!t]
\centering
%\vspace{-0.4cm}
\includegraphics[width=1\textwidth]{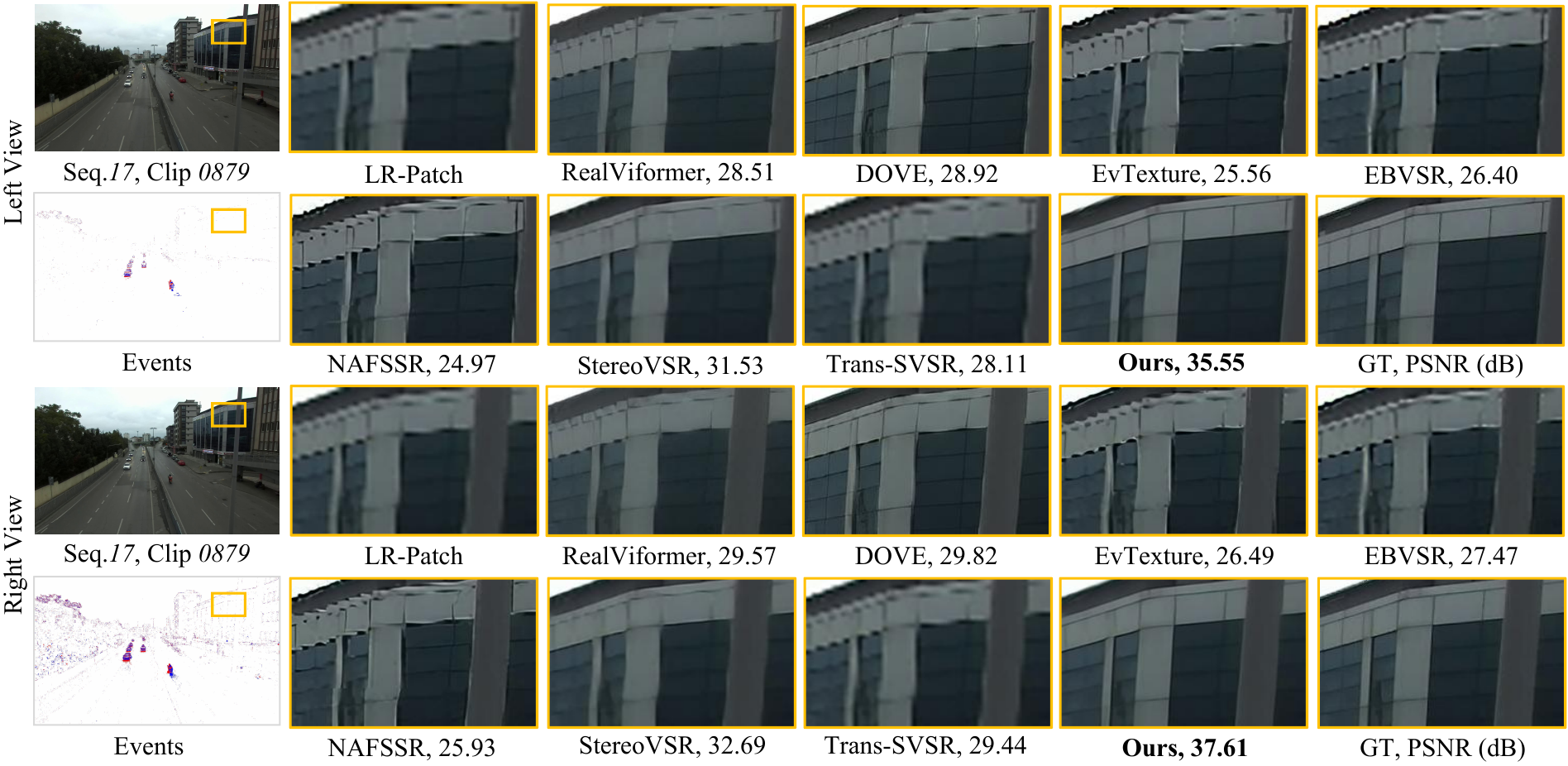}
% \vspace{-0.7cm}
\caption{Qualitative comparison of existing methods \cite{10.1007/978-3-031-73397-0_24, NEURIPS2025_7b04ec5f, kai2024evtexture, 10222922, 9857435, xu2021stereo, 9857357} for $\times 4$ stereo video super-resolution on the synthetic SVSR-Set. Best viewed at $400$\% zoom.
}
\label{fig:SVSR}
% \vspace{-0.25cm}
\end{figure*}

% \subsection{Experimental Results and Analysis}
\subsection{Quantitative Comparison and Analysis}

\noindent\textbf{1) Comparison with SOTAs}
\vspace{0.1cm}

Tab.~\ref{tab:main_results} presents quantitative comparisons of various methods for $\times4$ stereo video super-resolution on DSEC and SVSR-Set.
Our method consistently outperforms 
RGB-based VSR \cite{9879156,Li_Liu_Shang_Liu_Wan_Feng_2024,sun2026one,10.1007/978-3-031-73397-0_24,liu2025ultravsr,NEURIPS2025_7b04ec5f}, 
event-guided VSR \cite{kai2024evtexture,10222922}, 
RGB-based SISR \cite{dai2021feedback,9522905,9857435}, and 
RGB-based SVSR \cite{xu2021stereo,9857357} methods across all evaluation metrics on both datasets.
Specifically, our method achieves a PSNR of $37.37$ \textit{dB} and an SSIM of $0.947$ on DSEC, 
outperforming the strongest baseline StereoVSR~\cite{xu2021stereo}. 
Likewise, our method achieves a PSNR of $33.61$ \textit{dB} and an SSIM of $0.944$ on SVSR-Set, 
exceeding StereoVSR~\cite{xu2021stereo} by $2.08$ \textit{dB} and $0.02$, respectively.

Furthermore, our method achieves the lowest LPIPS and $E_{\mathrm{warp}}$ on both datasets, 
indicating better perceptual quality and temporal consistency.
Our method also achieves the lowest EPE of $1.376$ and $1.994$ on DSEC and SVSR-Set, respectively, denoting more accurate cross-view geometric correspondences.
These results validate the benefits of jointly exploiting event cues, temporal cues, and cross-view complementary cues to reconstruct stereo videos with faithful textures and consistent stereo geometry.

\vspace{0.1cm}
\noindent\textbf{2) Efficiency Analysis}
\vspace{0.1cm}

As presented in Tab.~\ref{tab:efficiency},
we evaluate the computational efficiency of representative methods in terms of parameter count and runtime. 
For a fair comparison, 
the runtime of single VSR methods is multiplied by $2$ to account for separate inference on both stereo views. 
Notably, SSRDE-FNet~\cite{dai2021feedback}, iPASSR~\cite{9522905}, and NAFSSR~\cite{9857435} only exploit spatial information without modeling inter-frame temporal dependencies. 
Although these methods incur relatively low computational costs, 
they achieve inferior reconstruction performance on stereo videos. 
In contrast, our method achieves a runtime of $290.4$ \textit{ms} with only $8.72$ \textit{M} parameters, outperforming several representative VSR and SVSR methods in computational efficiency.
This observation demonstrates that our method attains a desirable trade-off between reconstruction performance and computational overhead.

\begin{figure}[!t]
\centering
\includegraphics[width=0.48\textwidth]{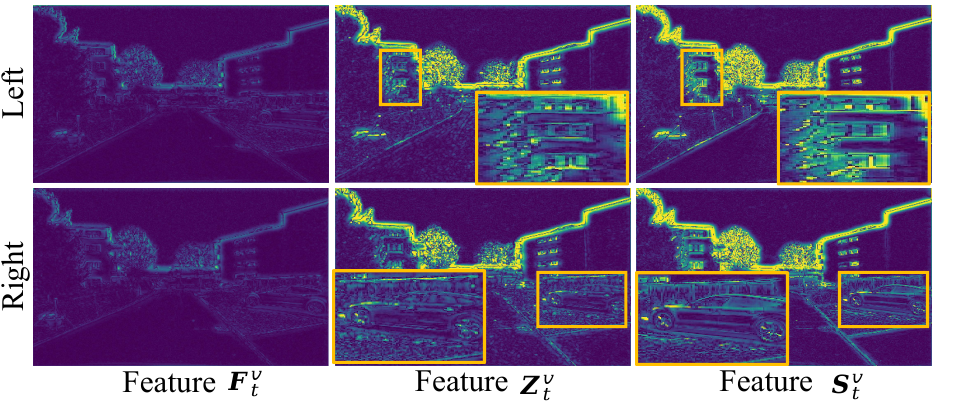}
% \vspace{-0.7cm}
\caption{
Feature visualization of outputs from the encoder, EGTE and CVTI modules for the left and right views. Best viewed at $400$\% zoom.
}
\label{fig:Feature}
% \vspace{-0.45cm}
\end{figure}

\subsection{Qualitative Comparison and Analysis}

\noindent\textbf{1) Comparison with SOTAs}
\vspace{0.1cm}

Figs.~\ref{fig:DSEC} and~\ref{fig:SVSR} present qualitative comparisons on the real-world DSEC and synthetic SVSR-Set, respectively.
Existing methods \cite{9879156,Li_Liu_Shang_Liu_Wan_Feng_2024,sun2026one,10.1007/978-3-031-73397-0_24,liu2025ultravsr,NEURIPS2025_7b04ec5f,kai2024evtexture,10222922,dai2021feedback,9522905,9857435,xu2021stereo,9857357}
frequently produce blurry textures, artifacts, and structural distortions within reconstructed outputs.
Concretely,
on DSEC,
these methods struggle to recover dense grid patterns on building facades, resulting in missing texture details and distorted structures.
On SVSR-Set, these methods often produce blurry edges and fail to preserve fine-grained structures around building boundaries and windows.
In contrast, our method reconstructs sharper textures and more accurate structures, delivering outputs that closely match the ground-truth. 

\begin{figure}[t!]
\centering
% \vspace{-0.1cm}
\includegraphics[width=0.9\linewidth]{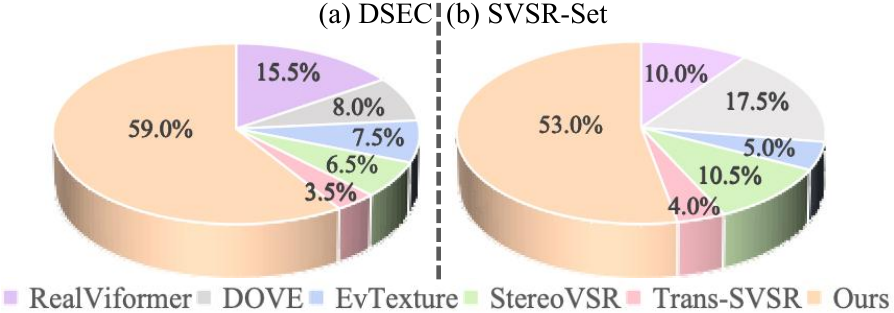}
% \vspace{-0.3cm}
\caption{User study on DSEC and SVSR-Set.}
\label{fig:User}
 \vspace{-0.3cm}
\end{figure}

\vspace{0.1cm}
\noindent\textbf {2) Feature Visualization}
\vspace{0.1cm}

% 重新写下这块内容  我觉这还蛮重要的 （你看下我写的可准确）
Fig.~\ref {fig:Feature} visualizes feature maps outputted from the encoder, the EGTE module and the CVTI module of the proposed EvSVSR for both left and right views.
Compared with the encoder features $\textbf{\textit{F}}_t^v$, the EGTE outputs $\textbf{\textit{Z}}_t^v$ 
show enhanced responses to texture and structural cues via injected event information, verifying the local enhancement capability of EGTE.
In contrast,
the CVTI outputs $\textbf{\textit{S}}_t^v$
further activate informative structural cues by exploiting cross-view complementary information.
These visualizations show that our method captures fine-grained texture cues and establishes reliable cross-view feature alignment.

\vspace{0.1cm}
\noindent\textbf{3) User Study}
\vspace{0.1cm}

To further evaluate the visual quality of our method, we conduct a user study. 
We compare our method with five competitive super-resolution methods, 
including RealViformer~\cite{10.1007/978-3-031-73397-0_24}, DOVE~\cite{NEURIPS2025_7b04ec5f}, EvTexture~\cite{kai2024evtexture}, StereoVSR~\cite{xu2021stereo}, and Trans-SVSR~\cite{9857357}. 
We select $5$ cases from each dataset and invite $40$ volunteers to participate in the study.
For each test case, we randomly shuffle the reconstruction outputs from all six methods and ask volunteers to select the result with the highest visual quality.
As indicated in Fig.~\ref{fig:User},
our method receives $59.0$\% and $53.0$\% of the votes on DSEC and SVSR-Set, respectively.
These results indicate a clear preference for our method over the competing methods and further validate its effectiveness in producing high-quality stereo video reconstructions.

\begin{table}[t]
\centering
\footnotesize
\renewcommand\tabcolsep{6pt}
% \vspace{-0.3cm}
\caption{Ablation study of the event signal and the proposed CVTI module.}
% \vspace{-0.3cm}
\label{tab:EGTE}
\resizebox{1\linewidth}{!}{
\begin{tabular}{ccccccc}
\toprule
\textbf{Event} & \textbf{CVTI} & \textit{PSNR}$\uparrow$ & \textit{SSIM}$\uparrow$ & \textit{LPIPS}$\downarrow$ & $E_{warp}\downarrow$ & \textit{EPE} $\downarrow$ \\
\midrule
$\times$ & $\checkmark$ & 35.67 & 0.932 & 0.281 & 0.009 & 1.556 \\
$\checkmark$ & $\times$ & 36.69 & 0.942 & 0.281 & 0.007 & 1.539 \\
\cellcolor{blue!10}$\checkmark$ & \cellcolor{blue!10}$\checkmark$ &\cellcolor{blue!10} \textbf{37.37} &\cellcolor{blue!10} \textbf{0.947} &\cellcolor{blue!10} \textbf{0.258} &\cellcolor{blue!10} \textbf{0.007} &\cellcolor{blue!10} \textbf{1.376} \\
\bottomrule
\end{tabular}
}
% \vspace{-0.3cm}
\end{table}
\begin{table}[t]
\centering
\footnotesize
\renewcommand\tabcolsep{6pt}
% \vspace{-0.3cm}
\caption{Generalization of the proposed CVTI module.}
% \vspace{-0.3cm}
\label{tab:cvtt}
\resizebox{1\linewidth}{!}{
\begin{tabular}{lccccc}
\toprule
\textbf{Method} & \textit{PSNR}$\uparrow$ & \textit{SSIM}$\uparrow$ & \textit{LPIPS}$\downarrow$ & $E_{warp}\downarrow$ & \textit{EPE} $\downarrow$ \\
\midrule
EvTexture & 28.75 & 0.849 & 0.387 & 0.045 & 2.990 \\
\cellcolor{blue!10}+CVTI &\cellcolor{blue!10} \textbf{36.63} &\cellcolor{blue!10} \textbf{0.941} &\cellcolor{blue!10} \textbf{0.279} &\cellcolor{blue!10} \textbf{0.007} &\cellcolor{blue!10} \textbf{1.510} \\
\midrule
EBVSR & 30.12 & 0.876 & 0.362 & 0.032 & 1.964 \\
\cellcolor{blue!10}+CVTI &\cellcolor{blue!10} \textbf{36.25} &\cellcolor{blue!10} \textbf{0.939} & \cellcolor{blue!10}\textbf{0.286} & \cellcolor{blue!10}\textbf{0.008} & \cellcolor{blue!10}\textbf{1.569} \\
\bottomrule
\end{tabular}
}
% \vspace{-0.3cm}
\end{table}

\subsection{Ablation Study}
\label{Ablation Study}

\begin{figure}[!t]
\centering
\includegraphics[width=0.48\textwidth]{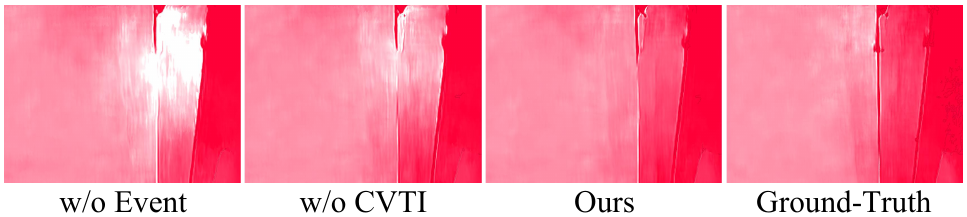}
% \vspace{-0.7cm}
\caption{
Parallax flow visualization with diverse configurations.
}
\label{fig:flow}
% \vspace{-0.45cm}
\end{figure}

\noindent\textbf{Role of Event Signal.}
As shown in the first row of Tab.~\ref{tab:EGTE}, 
removing the event signal from the EGTE module 
degrades performance across all evaluation metrics, demonstrating its importance for stereo video reconstruction. 
Beyond quantitative evaluation,
we further examine cross-view geometric consistency by visualizing the parallax flow between the left and right views.
As shown in the first column of Fig.~\ref{fig:flow}, 
the variant without event signal yields larger deviations from the ground truth.
These results indicate that event information contributes to both texture restoration and cross-view geometric consistency.

\begin{figure}[!t]
\centering
% \vspace{-0.1cm}
\includegraphics[width=0.48\textwidth]{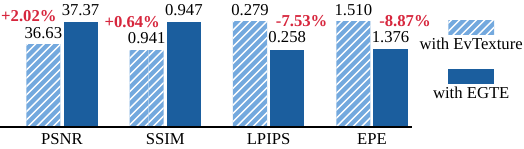}
% \vspace{-0.7cm}
\caption{
EvTexture~\cite{kai2024evtexture} versus
the proposed EGTE module.
}
\label{fig:evtexture_vs_egte}
% \vspace{-0.45cm}
\end{figure}

\noindent\textbf{Superiority of EGTE.}
We replace the proposed EGTE module with the event enhancement module from EvTexture~\cite{kai2024evtexture} for comparison.
As shown in Fig.~\ref{fig:evtexture_vs_egte}, 
our EGTE improves PSNR and SSIM by $2.02$\% and $0.64$\%, while reducing LPIPS and EPE by $7.53$\% and $8.87$\% compared with the replacement variant. 
These results demonstrate the superiority of the proposed EGTE.

\noindent\textbf{Effectiveness of CVTI.}
As shown in the second row of Tab.~\ref{tab:EGTE}, 
removing CVTI reduces PSNR from $37.37$ to $36.69$ \textit{dB}.
Meanwhile, EPE also increases from $1.376$ to $1.539$. 
These results demonstrate that CVTI improves both reconstruction quality and stereo consistency. 
To complement these quantitative results, 
we further examine cross-view geometric consistency by comparing parallax flows of reconstruction outputs in the second column of Fig.~\ref{fig:flow}.
Compared with the variant without CVTI, the full model produces parallax flows that are more consistent with the ground-truth.
This indicates that CVTI effectively integrates complementary information to strengthen cross-view geometric consistency.

\noindent\textbf{Generalization of CVTI.}
To assess the generalizability of CVTI, we embed it into two event-guided VSR methods, i.e.,
EvTexture~\cite{kai2024evtexture} and EBVSR~\cite{10222922}.
As shown in Tab.~\ref{tab:cvtt}, incorporating CVTI consistently improves the reconstruction quality of both methods. 
Specifically, CVTI increases the PSNR of EvTexture~\cite{kai2024evtexture} and EBVSR~\cite{10222922}, 
while improving SSIM and lowering LPIPS. 
Furthermore, both modified models achieve lower $E_{warp}$ and EPE, indicating enhanced cross-view consistency.
These results suggest that the designed CVTI module can be integrated into diverse event-guided VSR frameworks to boost stereo reconstruction quality and cross-view consistency.

\section{Conclusion}
In this paper, we present the first investigation of event information for SVSR and propose an event-guided SVSR framework. 
Our framework employs the EGTE module to recover high-frequency texture details in regions with complex motion and the CVTI module to exploit cross-view complementary information for better stereo consistency. 
Extensive experiments on real-world and synthetic datasets demonstrate that our method outperforms existing super-resolution methods in reconstruction quality and stereo consistency, while maintaining favorable computational efficiency. 
This work highlights the potential of event signals for stereo video restoration and provides new insights for future research.

%\section*{LLM Usage Statement}
%LLM was used to assist with language polishing, grammar correction, and improving the clarity of the manuscript. 
%All scientific ideas, methodological designs, experimental results, and conclusions were developed and verified by the authors.

{
    \small
    \bibliographystyle{ieeenat_fullname}
    \bibliography{main}

@article{10462902,
  author={Liang, Jingyun and Cao, Jiezhang and Fan, Yuchen and others},
  journal={IEEE Transactions on Image Processing (TIP)}, 
  title={VRT: A Video Restoration Transformer}, 
  year={2024},
  volume={33},
  number={},
  pages={2171--2182}
  }

@article{wen2022video,
  title={Video super-resolution via a spatio-temporal alignment network},
  author={Wen, Weilei and Ren, Wenqi and Shi, Yinghuan and others},
  journal={IEEE Transactions on Image Processing (TIP)},
  volume={31},
  pages={1761--1773},
  year={2022}
}

@inproceedings{9879776,
  author={Zhou, Kun and Li, Wenbo and Lu, Liying and others},
  booktitle={IEEE/CVF Conference on Computer Vision and Pattern Recognition (CVPR)}, 
  title={Revisiting Temporal Alignment for Video Restoration}, 
  year={2022},
  pages={6043--6052}
  }

@inproceedings{9879547,
  author={Chiche, Benjamin Naoto and Woiselle, Arnaud and Frontera-Pons, Joana and others},
  booktitle={IEEE/CVF Conference on Computer Vision and Pattern Recognition (CVPR)}, 
  title={Stable Long-Term Recurrent Video Super-Resolution}, 
  year={2022},
  pages={827--836}
  }

@inproceedings{11445549,
  author={Bernasconi, Michael and Djelouah, Abdelaziz and Zhang, Yang and others},
  booktitle={IEEE/CVF International Conference on Computer Vision (ICCV)}, 
  title={LDIP: Long Distance Information Propagation for Video Super-Resolution}, 
  year={2025},
  pages={11558--11567}
  }

@inproceedings{9577681,
  author={Chan, Kelvin C.K. and Wang, Xintao and Yu, Ke and others},
  booktitle={IEEE/CVF Conference on Computer Vision and Pattern Recognition (CVPR)}, 
  title={BasicVSR: The Search for Essential Components in Video Super-Resolution and Beyond}, 
  year={2021},
  pages={4945--4954}
  }

@inproceedings{9879156,
  author={Chan, Kelvin C.K. and Zhou, Shangchen and Xu, Xiangyu and others},
  booktitle={IEEE/CVF Conference on Computer Vision and Pattern Recognition (CVPR)}, 
  title={BasicVSR++: Improving Video Super-Resolution with Enhanced Propagation and Alignment}, 
  year={2022},
  pages={5962--5971}
  }

@inproceedings{10656731,
  author={Zhou, Xingyu and Zhang, Leheng and Zhao, Xiaorui and others},
  booktitle={IEEE/CVF Conference on Computer Vision and Pattern Recognition (CVPR)}, 
  title={Video Super-Resolution Transformer with Masked Inter\&Intra-Frame Attention}, 
  year={2024},
  pages={25399--25408}
  }

@inproceedings{chen2026stcdit,
  title={Stcdit: Spatio-temporally consistent diffusion transformer for high-quality video super-resolution},
  author={Chen, Junyang and Dong, Jiangxin and Sun, Long and others},
  booktitle={IEEE/CVF Conference on Computer Vision and Pattern Recognition (CVPR)},
  pages={38281--38290},
  year={2026}
}

@inproceedings{wang2025seedvr,
  title={Seedvr: Seeding infinity in diffusion transformer towards generic video restoration},
  author={Wang, Jianyi and Lin, Zhijie and Wei, Meng and others},
  booktitle={IEEE/CVF Conference on Computer Vision and Pattern Recognition (CVPR)},
  pages={2161--2172},
  year={2025}
}

@inproceedings{shang2024arbitrary,
  title={Arbitrary-scale video super-resolution with structural and textural priors},
  author={Shang, Wei and Ren, Dongwei and Zhang, Wanying and others},
  booktitle={European Conference on Computer Vision (ECCV)},
  pages={73--90},
  year={2024}
}

@article{xiao2026event,
  title={Event-Guided Online Video Super-Resolution},
  author={Xiao, Zeyu and Wang, Xinchao},
  journal={IEEE Transactions on Image Processing (TIP)},
  pages={8678--8690},
  volume={35}, 
  year={2026}
}

@inproceedings{Xiao_Wang_2026,
title={Exploiting Blurry Representations for Event-guided Video Super-Resolution}, 
volume={40}, 
number={13}, 
booktitle={AAAI Conference on Artificial Intelligence (AAAI)},
author={Xiao, Zeyu and Wang, Xinchao}, 
year={2026}, 
pages={11032--11041} 
}

@article{gallego2020event,
  title={Event-based vision: A survey},
  author={Gallego, Guillermo and Delbr{\"u}ck, Tobi and Orchard, Garrick and others},
  journal={IEEE Transactions on Pattern Analysis and Machine Intelligence (TPAMI)},
  volume={44},
  number={1},
  pages={154--180},
  year={2020}
}

@inproceedings{xiao2025event,
  title={Event-based video super-resolution via state space models},
  author={Xiao, Zeyu and Wang, Xinchao},
  booktitle={IEEE/CVF Conference on Computer Vision and Pattern Recognition (CVPR)},
  pages={12564--12574},
  year={2025}
}

@inproceedings{Kai_Zhang_Wang_Xiao_Xiong_Sun_2025, 
title={Event-Enhanced Blurry Video Super-Resolution}, 
volume={39}, 
number={4}, 
booktitle={AAAI Conference on Artificial Intelligence (AAAI)},
author={Kai, Dachun and Zhang, Yueyi and Wang, Jin and others},
year={2025},
pages={4175--4183} }

@inproceedings{9857357,
  author={Imani, Hassan and Islam, Md Baharul and Wong, Lai-Kuan},
  booktitle={IEEE/CVF Conference on Computer Vision and Pattern Recognition Workshops (CVPRW)}, 
  title={A New Dataset and Transformer for Stereoscopic Video Super-Resolution}, 
  year={2022},
  pages={705--714}
  }

@article{10440343,
  author={Hayat, Mansoor and Aramvith, Supavadee},
  journal={IEEE Access}, 
  title={E-SEVSR—Edge Guided Stereo Endoscopic Video Super-Resolution}, 
  year={2024},
  volume={12},
  number={},
  pages={30893--30906}
  }

@inproceedings{xu2021stereo,
  title={Stereo Video Super-Resolution via Exploiting View-Temporal Correlations},
  author={Xu, Ruikang and Xiao, Zeyu and Yao, Mingde and others},
  booktitle={ACM International Conference on Multimedia (ACMMM)},
  pages={460--468},
  year={2021}
}

@inproceedings{zhuang2026flashvsr,
  title={Flashvsr: Towards real-time diffusion-based streaming video super resolution},
  author={Zhuang, Junhao and Guo, Shi and Cai, Xin and others},
  booktitle={IEEE/CVF Conference on Computer Vision and Pattern Recognition (CVPR)},
  pages={43482--43493},
  year={2026}
}

@article{zhu2026fcvsr,
  title={Fcvsr: A frequency-aware method for compressed video super-resolution},
  author={Zhu, Qiang and Zhang, Fan and Chen, Feiyu and others},
  journal={IEEE Transactions on Multimedia (TMM)},
  volume={28},
  pages={1--14},
  year={2026}
}

@inproceedings{sun2026one,
  title={One-step diffusion for detail-rich and temporally consistent video super-resolution},
  author={Sun, Yujing and Sun, Lingchen and Liu, Shuaizheng and others},
  booktitle={Advances in Neural Information Processing Systems (NeurIPS)},
  volume={38},
  pages={172821--172841},
  year={2026}
}

@inproceedings{NEURIPS2025_7b04ec5f,
 author = {Chen, Zheng and Zou, Zichen and Zhang, Kewei and others},
 booktitle = {Advances in Neural Information Processing Systems (NeurIPS)},
 pages = {85218--85237},
 title = {DOVE: Efficient One-Step Diffusion Model for Real-World Video Super-Resolution},
 year = {2025}
}

@article{11417318,
  author={Li, Hao and Chen, Xiang and Dong, Jiangxin and others},
  journal={IEEE Transactions on Pattern Analysis and Machine Intelligence (TPAMI)}, 
  title={Collaborative Feedback Discriminative Propagation for Video Super-Resolution}, 
  year={2026},
  volume={48},
  number={7},
  pages={7835--7845}
  }

@inproceedings{10658213,
  author={Yang, Yixin and Liang, Jinxiu and Yu, Bohan and others},
  booktitle={IEEE/CVF Conference on Computer Vision and Pattern Recognition (CVPR)}, 
  title={Latency Correction for Event-Guided Deblurring and Frame Interpolation}, 
  year={2024},
  pages={24977--24986}}

@inproceedings{11093199,
  author={Chen, Jingxi and Feng, Brandon Y. and Cai, Haoming and others},
  booktitle={IEEE/CVF Conference on Computer Vision and Pattern Recognition (CVPR)}, 
  title={Repurposing Pre-trained Video Diffusion Models for Event-based Video Interpolation}, 
  year={2025},
  pages={12456--12466}}

@article{11614017,
  author={Liu, Yuhan and Deng, Yongjian and Fu, Linghui and others},
  journal={IEEE Transactions on Circuits and Systems for Video Technology (TCSVT)}, 
  title={Video Frame Interpolation via Asymmetric Refinement With the Event-Based Reference}, 
  year={2026},
  volume={36},
  number={11},
  pages={16789--16804}
  }

@inproceedings{11444320,
  author={Zhang, Pengjie and Zhu, Lin and Wang, Xiao and others},
  booktitle={IEEE/CVF International Conference on Computer Vision (ICCV)}, 
  title={EMatch: A Unified Framework for Event-Based Optical Flow and Stereo Matching}, 
  year={2025},
  pages={5845--5855}
  }

@inproceedings{11092933,
  author={Zhou, Hanyu and Wang, Haonan and Liu, Haoyue and others},
  booktitle={IEEE/CVF Conference on Computer Vision and Pattern Recognition (CVPR)}, 
  title={Bridge Frame and Event: Common Spatiotemporal Fusion for High-Dynamic Scene Optical Flow}, 
  year={2025},
  pages={27904--27913}
  }

@article{10576683,
  author={Liu, Minjie and Wang, Hongjian and Yoon, Kuk-Jin and others},
  journal={IEEE Transactions on Artificial Intelligence (TAI)}, 
  title={Disentangled Cross-modal Fusion for Event-Guided Image Super-resolution}, 
  year={2024},
  volume={5},
  number={10},
  pages={5314--5324}
  }

@article{10994378,
  author={Duan, Peiqi and Li, Boyu and Yang, Yixin and others},
  journal={IEEE Transactions on Pattern Analysis and Machine Intelligence (TPAMI)}, 
  title={EventAid: Benchmarking Event-Aided Image/Video Enhancement Algorithms With Real-Captured Hybrid Dataset}, 
  year={2025},
  volume={47},
  number={8},
  pages={6959--6973}
  }

@inproceedings{11093954,
  author={Wei, Shuoyan and Li, Feng and Tang, Shengeng and others},
  booktitle={IEEE/CVF Conference on Computer Vision and Pattern Recognition (CVPR)}, 
  title={EvEnhancer: Empowering Effectiveness, Efficiency and Generalizability for Continuous Space-Time Video Super-Resolution with Events}, 
  year={2025},
  pages={17755--17766}
  }

@inproceedings{pmlr-v87-rebecq18a,
  title={ESIM: an Open Event Camera Simulator},
  author={Rebecq, Henri and Gehrig, Daniel and Scaramuzza, Davide},
  booktitle={Conference on Robot Learning (CoRL)},
  pages={969--982},
  year={2018},
  volume={87}
}

@article{11369964,
  author={Kai, Dachun and Lu, Jiayao and Zhang, Yueyi and others},
  journal={IEEE Transactions on Pattern Analysis and Machine Intelligence (TPAMI)}, 
  title={EvTexture++: Event-Driven Texture Enhancement for Video Super-Resolution}, 
  year={2026},
  volume={48},
  number={6},
  pages={6642--6659}
  }

@article{9387069,
  author={Gehrig, Mathias and Aarents, Willem and Gehrig, Daniel and others},
  journal={IEEE Robotics and Automation Letters (RAL)}, 
  title={DSEC: A Stereo Event Camera Dataset for Driving Scenarios}, 
  year={2021},
  volume={6},
  number={3},
  pages={4947--4954}
  }

@inproceedings{Li_Liu_Shang_Liu_Wan_Feng_2024, 
title={SAVSR: Arbitrary-Scale Video Super-Resolution via a Learned Scale-Adaptive Network}, 
volume={38}, 
number={4}, 
booktitle={AAAI Conference on Artificial Intelligence (AAAI)}, 
author={Li, Zekun and Liu, Hongying and Shang, Fanhua and others}, 
year={2024}, 
month={Mar.},
pages={3288--3296} 
}

@inproceedings{10.1007/978-3-031-73397-0_24,
  title={RealViformer: Investigating Attention for Real-World Video Super-Resolution},
  author={Zhang, Yuehan and Yao, Angela},
  booktitle={European Conference on Computer Vision (ECCV)},
  year={2024},
  pages={412--428}
}

@inproceedings{10222922,
  author={Kai, Dachun and Zhang, Yueyi and Sun, Xiaoyan},
  booktitle={IEEE International Conference on Image Processing (ICIP)}, 
  title={Video Super-Resolution Via Event-Driven Temporal Alignment}, 
  year={2023},
  pages={2950--2954}
  }

@inproceedings{9522905,
  author={Wang, Yingqian and Ying, Xinyi and Wang, Longguang and others},
  booktitle={IEEE/CVF Conference on Computer Vision and Pattern Recognition Workshops (CVPRW)}, 
  title={Symmetric Parallax Attention for Stereo Image Super-Resolution}, 
  year={2021},
  pages={766--775}
  }

@inproceedings{9857435,
  author={Chu, Xiaojie and Chen, Liangyu and Yu, Wenqing},
  booktitle={IEEE/CVF Conference on Computer Vision and Pattern Recognition Workshops (CVPRW)}, 
  title={NAFSSR: Stereo Image Super-Resolution Using NAFNet}, 
  year={2022},
  pages={1238--1247}
  }

@inproceedings{kai2024evtexture,
  title={{E}v{T}exture: {E}vent-driven {T}exture {E}nhancement for {V}ideo {S}uper-{R}esolution},
  author={Kai, Dachun and Lu, Jiayao and Zhang, Yueyi and others},
  booktitle={International Conference on Machine Learning (ICML)},
  pages={22817--22839},
  year={2024}
}

@inproceedings{liu2025ultravsr,
  title={UltraVSR: Achieving Ultra-Realistic Video Super-Resolution with Efficient One-Step Diffusion Space},
  author={Liu, Yong and Pan, Jinshan and Li, Yinchuan and others},
  booktitle={ACM International Conference on Multimedia (ACMMM)},
  year={2025},
  Pages= {7785--7794}
}

@inproceedings{dai2021feedback,
  title={Feedback Network for Mutually Boosted Stereo Image Super-Resolution and Disparity Estimation},
  author={Dai, Qinyan and Li, Juncheng and Yi, Qiaosi and others},
  booktitle={ACM International Conference on Multimedia (ACMMM)},
  pages={1985--1993},
  year={2021}
}

@inproceedings{10205204,
  author={Wu, Zhiliang and Sun, Changchang and Xuan, Hanyu and others},
  booktitle={IEEE/CVF Conference on Computer Vision and Pattern Recognition (CVPR)}, 
  title={Deep Stereo Video Inpainting}, 
  year={2023},
  pages={5693--5702}
  }

@inproceedings{Wu_Li_Xu_Fan_Yang_2026, 
title={DLVINet: Advancing Dual-Lens Video Inpainting Beyond Parallax Constraints}, 
volume={40}, 
number={13},
booktitle={AAAI Conference on Artificial Intelligence (AAAI)}, 
author={Wu, Zhiliang and Li, Kun and Xu, Yunqiu and others}, 
year={2026}, 
pages={10888--10896} 
}

@inproceedings{11092792,
  author={Xu, Yiran and Park, Taesung and Zhang, Richard and others},
  booktitle={IEEE/CVF Conference on Computer Vision and Pattern Recognition (CVPR)}, 
  title={VideoGigaGAN: Towards Detail-rich Video Super-Resolution}, 
  year={2025},
  pages={2139--2149}
  }

@article{1284395,
  author={Zhou Wang and Bovik,  Alan C. and Sheikh, Hamid R. and others},
  journal={IEEE Transactions on Image Processing (TIP)}, 
  title={Image quality assessment: from error visibility to structural similarity}, 
  year={2004},
  volume={13},
  number={4},
  pages={600--612}
  }

@inproceedings{8578166,
  author={Zhang, Richard and Isola, Phillip and Efros, Alexei A. and others},
  booktitle={IEEE/CVF Conference on Computer Vision and Pattern Recognition (CVPR)}, 
  title={The Unreasonable Effectiveness of Deep Features as a Perceptual Metric}, 
  year={2018},
  pages={586--595}
  }

@inproceedings{8099774,
  author={Ranjan, Anurag and Black, Michael J.},
  booktitle={IEEE/CVF Conference on Computer Vision and Pattern Recognition (CVPR)}, 
  title={Optical Flow Estimation Using a Spatial Pyramid Network}, 
  year={2017},
  pages={2720--2729}
  }

@inproceedings{8100101,
  author={Lai, Wei-Sheng and Huang, Jia-Bin and Ahuja, Narendra and others},
  booktitle={IEEE/CVF Conference on Computer Vision and Pattern Recognition (CVPR)}, 
  title={Deep Laplacian Pyramid Networks for Fast and Accurate Super-Resolution}, 
  year={2017},
  pages={5835--5843}
  }

@inproceedings{ballas2015delving,
  title={Delving deeper into convolutional networks for learning video representations},
  author={Ballas, Nicolas and Yao, Li and Pal, Chris and others},
  booktitle={International Conference on Learning Representations (ICLR)},
  year={2016},
  pages={1--11}
}

@inproceedings{9025464,
  author={Wang, Xintao and Chan, Kelvin C.K. and Yu, Ke and others},
  booktitle={IEEE/CVF Conference on Computer Vision and Pattern Recognition Workshops (CVPRW)}, 
  title={EDVR: Video Restoration With Enhanced Deformable Convolutional Networks}, 
  year={2019},
  pages={1954--1963}
  }

@inproceedings{shi2016real,
  title={Real-time single image and video super-resolution using an efficient sub-pixel convolutional neural network},
  author={Shi, Wenzhe and Caballero, Jose and Husz{\'a}r, Ferenc and others},
  booktitle={IEEE/CVF Conference on Computer Vision and Pattern Recognition (CVPR)},
  pages={1874--1883},
  year={2016}
}

@inproceedings{ronneberger2015u,
  title={U-net: Convolutional networks for biomedical image segmentation},
  author={Ronneberger, Olaf and Fischer, Philipp and Brox, Thomas},
  booktitle={International Conference on Medical Image Computing and Computer-Assisted Intervention (MICCAI)},
  pages={234--241},
  year={2015}
}
}


\end{document}